\documentclass[letterpaper, 10pt, conference]{ieeeconf}
\usepackage{float}
\usepackage{placeins}
\usepackage{booktabs}
\usepackage{graphicx}
\usepackage{amsmath,amssymb}
\usepackage{algorithm}
\usepackage{algpseudocode}
\usepackage{mdwlist}

\usepackage{textcomp}
\usepackage{cite}
\usepackage{url}

\usepackage{geometry}
\title{\LARGE \bf DeCoNav: Dialog enhanced Long-Horizon Collaborative Vision-Language Navigation}

\author{
\authorblockN{
Sunyao Zhou$^{1,2}$\authorrefmark{2},
Yunzi Wu$^{1,3}$\authorrefmark{2},
Tianhang Wang$^{1,3}$,
Xinhai Li$^{1}$\authorrefmark{3}\\
Guang Chen$^{3}$,
Lizheng Liu$^{2}$\authorrefmark{1},
Chenjia Bai$^{1}$,
Xuelong Li$^{1}$\authorrefmark{1}
}
\authorblockA{
\!$^{1}$Institute of Artificial Intelligence (TeleAI), China Telecom \hspace{1em}
$^{2}$Fudan University \hspace{1em}
$^{3}$Tongji University\\
\authorrefmark{2}Equal contribution \hspace{1em}
\authorrefmark{3}Project leader \hspace{1em}
\authorrefmark{1}Corresponding authors
}
}

\begin{document}
\maketitle
\thispagestyle{empty}
\pagestyle{empty}

\begin{abstract}
  Long-horizon collaborative vision-language navigation (VLN) is critical for multi-robot systems to accomplish complex tasks beyond the capability of a single agent. CoNavBench takes a first step by introducing the first collaborative long-horizon VLN benchmark with relay-style multi-robot tasks, a collaboration taxonomy, along with graph-grounded generation and evaluation to model handoffs and rendezvous in shared environments. However, existing benchmarks and evaluations often do not enforce strictly synchronized dual-robot rollout on a shared world timeline, and they typically rely on static coordination policies that cannot adapt when new cross-agent evidence emerges. We present \textbf{Dialog enhanced Long-Horizon Collaborative Vision-Language Navigation (DeCoNav)}, a decentralized framework that couples \textbf{event-triggered dialogue} with \textbf{dynamic task allocation and replanning} for real-time, adaptive coordination. In DeCoNav, robots exchange compact semantic states via dialogue without a central controller. When informative events such as new evidence, uncertainty, or conflicts arise, dialogue is triggered to dynamically reassign subgoals and replan under synchronized execution. Implemented in \textbf{DeCoNavBench} with 1,213 tasks across 176 HM3D scenes, DeCoNav improves the both-success rate (BSR) by 69.2\%, demonstrating the effectiveness of dialogue-driven, dynamically reallocated planning for multi-robot collaboration.

\end{abstract}

\begin{figure*}[t]
  \centering
  \includegraphics[width=\textwidth]{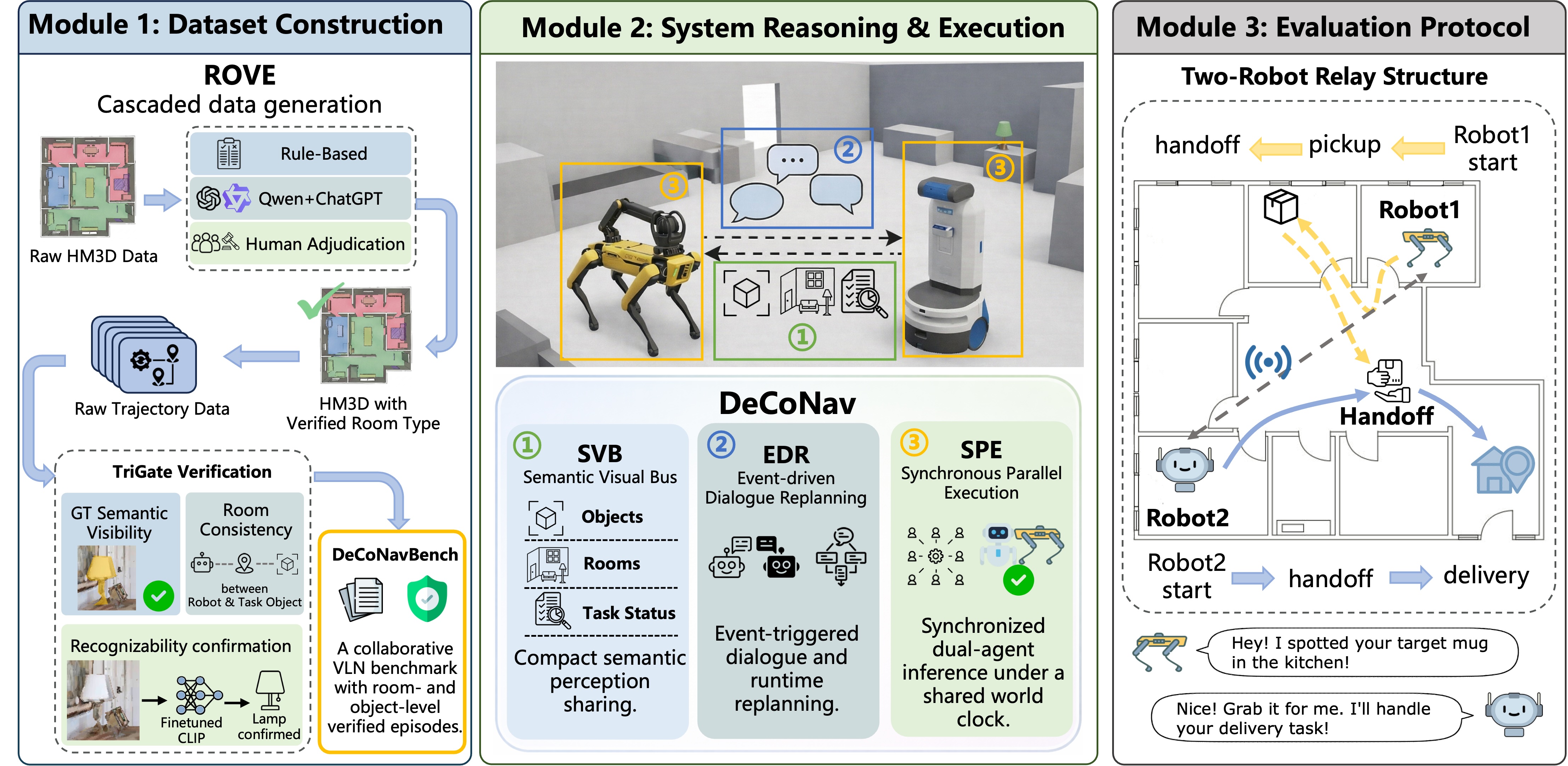}
  \caption{Overview of DeCoNav and DeCoNavBench. 
  \textbf{Module~1:} The ROVE pipeline constructs verified episodes 
  through rule-based inference, VLM classification, human adjudication 
  (RTSA), and triple-gate target verification (TriGate). 
  \textbf{Module~2:} DeCoNav coordinates dual-robot execution via three 
  coupled components: Semantic Visual Bus (SVB) for compact state 
  exchange, Event-driven Dialogue Replanning (EDR) for online 
  subtask reassignment, and Synchronous Parallel Execution (SPE) 
  for shared-clock inference. 
  \textbf{Module~3:} Two robots execute a relay task concurrently, 
  where Robot\,1 follows a start$\rightarrow$pickup$\rightarrow$handoff 
  trajectory and Robot\,2 follows a 
  start$\rightarrow$handoff$\rightarrow$delivery trajectory, 
  communicating via event-triggered dialogue.}
  \label{fig:teaser}
\end{figure*}

\section{Introduction}

Long-horizon navigation poses a fundamental challenge for embodied agents, demanding persistent memory, multi-stage reasoning, and robust error recovery across extended trajectories. Even a single agent struggles to meet these requirements in realistic indoor scenes. Collaborative VLN further compounds the challenge: multiple robots must interpret shared language instructions, synchronize their actions, and cope with individually limited, partial observations.

CoNavBench is a major step in this direction: it introduces relay-oriented collaborative tasks and a unified evaluation setting, making long-horizon collaborative VLN benchmarkable at scale~\cite{wang2026conavbench}. This directly addresses the earlier gap where collaborative VLN lacked a standardized task-and-metric framework.
Yet two limitations remain critical for system progress. First, collaborative quality is often not measured with truly synchronized dual-robot rollout in one shared world timeline, so many interaction effects are under-estimated or missed. Second, coordination strategies are typically static, relying on fixed role scheduling or predetermined plans. This limits their adaptability when robots encounter new evidence online, such as one robot observing a target early, receiving delayed partner messages, or requiring runtime route reassignment.

This paper addresses these limitations through a unified \emph{method-data} design in which policy design and benchmark construction are co-developed for the same synchronized collaborative setting.

At the method level, \textbf{Dialog enhanced Long-Horizon Collaborative Vision-Language Navigation (DeCoNav)} addresses both bottlenecks jointly. Each robot performs local reasoning but communicates compact semantic state rather than raw perception, preserving coordination signals under realistic communication constraints. When newly observed cross-agent evidence invalidates the current plan, the robot updates its policy context online to reassign relay behavior and navigation priorities. Crucially, both robots act in lockstep on a shared world clock, so the effect of communication and replanning is measured under true interaction rather than post-hoc aggregation of asynchronous rollouts. This formulation is deployment-consistent: in real two-machine setups, the same decentralized logic runs per robot, and only semantic messages are exchanged asynchronously under a freshness constraint. Concretely, this design consists of three components: \textbf{Semantic Visual Bus (SVB)} for semantic-state exchange, \textbf{Event-driven Dialogue Replanning (EDR)} for online policy revision, and \textbf{Synchronous Parallel Execution (SPE)} for strict dual-robot synchronized rollout.

At the data level, reliable evaluation requires episodes with correct room semantics and verifiable target observability. Otherwise, policy improvements can be confounded by annotation noise. \textbf{ROVE} (Room-and-Object Verified Episode Generation) provides a cascaded verification pipeline: \textbf{RTSA} enforces room-type semantic correctness for pickup, handoff, and delivery endpoints, while \textbf{TriGate} validates target stops through semantic visibility, room consistency, and visual recognizability checks. Using ROVE, CoNavBench is extended into \textbf{DeCoNavBench}, a benchmark designed for synchronized dual-robot evaluation and dynamic coordination analysis. This work makes the following contributions:
\begin{itemize}
  \item \textbf{DeCoNav:} a decentralized collaborative VLN framework that combines semantic-state communication, event-triggered online replanning, and synchronized dual-robot execution in a single closed-loop policy, with an algorithmically consistent extension to two-machine distributed deployment.
  \item \textbf{ROVE:} a room-and-object verified episode generation pipeline that improves benchmark reliability through cascaded semantic checks and adaptive stop-level repair.
  \item \textbf{DeCoNavBench:} a synchronized dual-robot extension of CoNavBench, built with ROVE, for reproducible evaluation of dynamic collaborative coordination.
\end{itemize}

\section{Related Work}
\subsection{Long-Horizon VLN}
Long-horizon VLN has progressed from short instruction following to multi-stage decision making with persistent memory and recovery. R2R first established path-following from natural language in indoor navigation~\cite{Anderson_2018_CVPR}, and VLN-CE subsequently moved evaluation to continuous control with embodied dynamics~\cite{krantz2020beyond}. To further increase task difficulty, R4R extended path length and alignment complexity~\cite{jain-etal-2019-stay}, while RxR broadened the scope to multilingual and fine-grained grounding supervision~\cite{ku2020room}. Beyond purely instructional settings, CVDN demonstrated that dialogue can reduce ambiguity in hard navigation states~\cite{thomason2020vision}. Underpinning these advances, Habitat provided a standard embodied simulation platform~\cite{Savva_2019_ICCV}, and HM3D further improved scene scale and diversity for robust evaluation~\cite{ramakrishnan2021habitat}.

On the evaluation side, nDTW/SDTW emphasized trajectory faithfulness beyond final success~\cite{49206}, and SCT added a dynamics-aware efficiency perspective~\cite{9636743}. More recent benchmarks have pushed toward increasingly challenging settings: IVLN highlighted iterative long-horizon interaction and memory effects~\cite{Krantz_2023_CVPR}, GOAT-Bench targeted multimodal lifelong-style navigation~\cite{Khanna_2024_CVPR}, and LHPR-VLN further stressed long-horizon planning and reasoning~\cite{song2025towards}.

On the model side, LM-Nav first connected large pretrained models to embodied navigation control~\cite{shah2023lm}. Subsequent work expanded this direction: InstructNav advanced zero-shot instruction following in unseen scenes~\cite{long2025instructnav}, Hi Robot pushed hierarchical VLA-style control for open-ended tasks~\cite{shi2025hi}, and Visual Language Maps emphasized map-level semantic grounding~\cite{10160969}. More recently, RL scaling on navigation foundation models improved interactive adaptation and robustness~\cite{he2025seeingexperiencingscalingnavigation}. Related goal formulations also evolved through image-goal and object-goal navigation~\cite{fried2018speaker}, and recent zero-shot object navigation methods further strengthened visual grounding with robust exploration interfaces~\cite{yadav2023ovrlv2simplestateofartbaseline,10610499,zhong2025topvnavunlockingtopviewspatial}. Despite this progress, most long-horizon protocols remain effectively single-robot and do not evaluate synchronized relay collaboration.

\subsection{Cooperative Multi-Robot VLN Benchmarks}
Multi-robot VLN extends single-agent pipelines by introducing coordination, communication, and shared-task execution. Early efforts explored diverse coordination paradigms: Co-NavGPT introduced VLM-driven cooperative frontier planning under decentralized perception~\cite{11302789}, MCoCoNav emphasized multimodal collaboration with chain-of-thought style coordination signals~\cite{Shen_Luo_Chen_Lv_Li_2025}, and CAMON focused on conversation-based cooperation for multi-object navigation~\cite{wu2024camoncooperativeagentsmultiobject}. Beyond task-specific systems, visual-language multi-agent planning studied planner-free coordination among multiple embodied agents~\cite{Brienza_2024}, and language-conditioned offline RL has been explored for multi-robot navigation policy learning~\cite{11127288}. Most notably, CoNavBench established a benchmarked setting for collaborative long-horizon relay tasks~\cite{wang2026conavbench}. However, the remaining gap is not basic decentralization itself, but reliable evaluation of synchronized interaction and online role reassignment when new cross-agent evidence appears.

\subsection{Visual Representations with Dialogue-Based Collaboration}

Scalable collaboration depends on compact, shareable semantic representations. CLIP provided a foundational semantic interface for object-level grounding~\cite{radford2021learning}, and EVA-CLIP further improved recognition robustness for large-scale visual-language alignment~\cite{sun2023evaclipimprovedtrainingtechniques}. Complementing these representation advances, semantic-agnostic and spatial-aware representation learning improved transfer in embodied navigation~\cite{10114576}. Building on stronger representations, recent pipelines increasingly combine semantic maps with model-based planning to stabilize long-horizon behavior~\cite{cheng2025navilaleggedrobotvisionlanguageaction,xu2024mobility}, and prospective exploration frameworks further improved decision quality in sparse-feedback settings~\cite{xue2026omninavunifiedframeworkprospective,11358651}.

On the communication side, dialogue-conditioned navigation has shown that language exchange is most useful when uncertainty is high and partner evidence is complementary~\cite{thomason2020vision}, a finding further supported by conversation-based multi-agent systems in cooperative settings~\cite{wu2024camoncooperativeagentsmultiobject}. DeCoNav builds on these lines with event-triggered replanning and strict synchronized execution, and DeCoNavBench is designed to evaluate this setting.

\section{DeCoNavBench Platform and ROVE Generation}
\label{sec:platform}

DeCoNavBench is designed to evaluate synchronized dual-robot collaborative navigation while rigorously controlling for semantic noise in the evaluation pipeline.


\subsection{ROVE Pipeline Overview}
\label{sec:rove}

DeCoNavBench employs \textbf{ROVE}, a cascaded verification pipeline that enables efficient episode generation through stop-level adaptive resampling. ROVE comprises two core modules: \textbf{RTSA} (Room-Type Semantic Alignment) and \textbf{TriGate} (Triple-Gate Target Verification), which together ensure that each endpoint candidate satisfies both semantic and observability constraints.

\subsection{Task Definition}
\label{sec:task_def}

We formulate collaborative VLN as a concurrent two-robot relay task within HM3D environments, defined as $\mathcal{T} = (\mathcal{E}, \mathcal{I}, R^{\text{FH}}, R^{\text{SH}})$, where $\mathcal{E}$ denotes the HM3D scene. Given a joint natural-language instruction $\mathcal{I}$, the first-half robot $R^{\text{FH}}$ executes the \emph{start$\rightarrow$pickup$\rightarrow$handoff} trajectory, while the second-half robot $R^{\text{SH}}$ performs the \emph{start$\rightarrow$handoff$\rightarrow$delivery} trajectory. Both robots operate simultaneously, receiving RGB-D observations and selecting from a discrete action space. An episode is deemed globally successful only if both robots complete all constituent subtasks along their respective trajectories. This relay structure naturally extends to larger scenes, where additional robots can be incorporated to handle more subtasks and collaboratively accomplish increasingly complex missions.

\subsection{Room-Type Semantic Alignment (RTSA)}
\label{sec:room_semantic}

RTSA addresses the absence of room-level semantic annotations in HM3D by assigning functional room-type labels to each region through a three-stage pipeline. First, a rule-based classifier leverages the semantic object inventory within each region to identify rooms with distinctive object signatures. Second, panoramic images rendered from each region centroid are fed to vision-language models to leverage the most recent advancements in the field. Specifically, we utilize the latest GPT-5.2 and Qwen3-VL-235B-A22B-Instruct~\cite{bai2025qwen3vltechnicalreport} for independent room-type classification. Finally, all assignments undergo manual review, with particular attention to disputed cases. The resulting verified labels are then incorporated into natural-language instructions during task generation, while also provide room-level progress signals for online evaluation (Sec.~\ref{subsec:room_recon_eval}).

\subsection{TriGate Target Verification}
\label{sec:triple_verify}

\begin{figure}[!t]
\centering
\includegraphics[width=0.95\columnwidth]{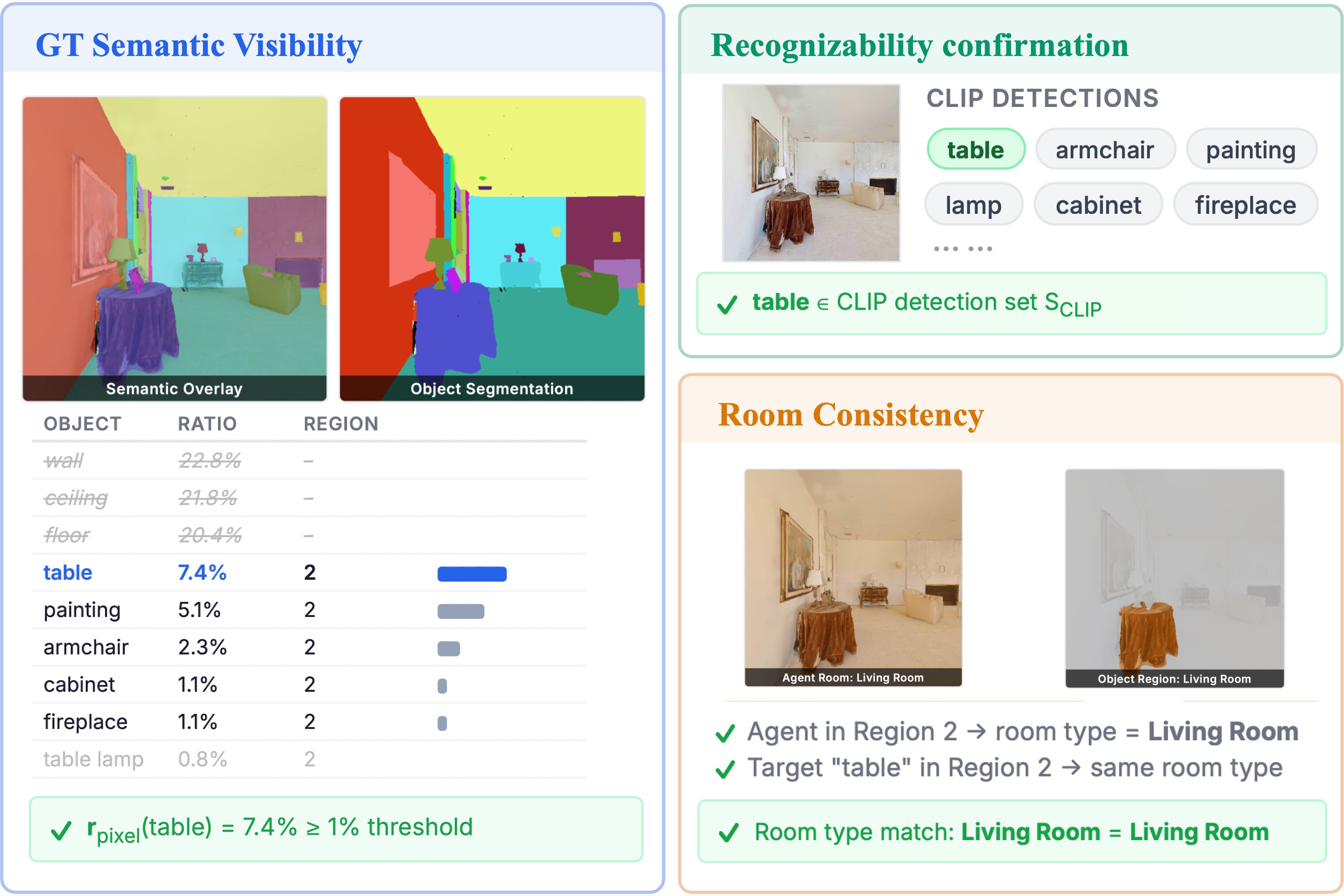}
\caption{Overview of the TriGate target verification pipeline. Each candidate waypoint must pass three gates: GT semantic visibility, room consistency, and CLIP-based recognizability confirmation.}
\label{fig:triple_verify}
\end{figure}

TriGate enforces three necessary conditions on each candidate waypoint through offline verification: \textbf{(a)}~GT semantic visibility, \textbf{(b)}~room consistency between the object and its surrounding region, and \textbf{(c)}~recognizability confirmation assessed by an HM3D-finetuned EVA-CLIP~\cite{sun2023evaclipimprovedtrainingtechniques}. Figure~\ref{fig:triple_verify} provides an overview of the TriGate target verification pipeline. 

A detailed analysis of dataset statistics is provided in Section~\ref{sec:data_statistics}.
With this reliability-first benchmark foundation in place, we next
introduce DeCoNav as the execution-time coordination framework built on
top of DeCoNavBench.

\section{DeCoNav Method}
\label{sec:collab_method}
DeCoNav is a decentralized framework for long-horizon collaborative VLN. Built on DeCoNavBench, this section focuses on how robots coordinate online through semantic communication and event-triggered dialogue replanning during execution. The framework operates as a closed-loop pipeline: at each global step, every robot processes its local observations, publishes a compact semantic state, evaluates whether new evidence warrants dialogue-triggered replanning, and then both robots execute actions synchronously.

The method uses three coupled modules: \textbf{Semantic Visual Bus (SVB)} for compact state exchange, \textbf{Event-driven Dialogue Replanning (EDR)} for adaptive goal update, and \textbf{Synchronous Parallel Execution (SPE)} for strict shared-time interaction. All coordination emerges from semantic messages and synchronized execution, with no centralized control between robots.

Rather than maintaining continuous communication, dialogue is triggered only by informative events. Typical triggers include stage completion, first reliable target discovery, conflict between local and partner evidence, or prolonged stagnation. After a trigger, robots exchange concise semantic dialogue, update subgoal priority, and continue the next synchronized step.

\subsection{Semantic Visual Bus (SVB)}
\label{subsec:shared_visual_representation}
SVB is the communication backbone of DeCoNav. Instead of exchanging raw
images or hidden states, each robot publishes a compact semantic packet
that includes current room, key observed objects, task-stage status, and
timestamp. This design keeps bandwidth low while preserving
coordination-critical information.

The shared bus stores per-robot state and lightweight coordination
memory:
\begin{equation}
  B_t=\{S_t^1,S_t^2,\mathcal{A}_t,\mathcal{R}_t,\mathcal{D}_t\},
\end{equation}
where $S_t^r$ denotes the semantic state of robot $r$ at step $t$
(including current room, observed objects, and task-stage status),
$\mathcal{A}_t$ is the anchor memory that records key object and
location evidence, $\mathcal{R}_t$ tracks the current role and
relay assignment for each robot, and $\mathcal{D}_t$ stores the
accumulated dialogue event history.

At each step, both robots write new semantic observations to $B_t$, and
the context composer injects valid partner hints into each robot's
runtime prompt. If partner evidence is missing or stale, control falls
back to local semantics, which preserves robustness under intermittent
communication.

\subsection{Event-driven Dialogue Replanning (EDR)}
\label{subsec:dialog_trigger_and_reassign}
EDR converts semantic events into online policy updates. Each robot
maintains current goal order, event history, and anchor memory. When a
trigger arrives, the robot rewrites the runtime instruction and updates
subgoal priority.


EDR follows three operations in each step: event extraction, event
filtering, and policy-context rewrite. In practice, this means that once
new evidence appears, both robots can re-rank navigation cues and adjust
relay responsibility without any central coordinator.

\subsection{Synchronous Parallel Execution (SPE)}
\label{subsec:execution_pipeline}
SPE enforces a shared world clock so that both robots act under the same causal timeline. At each step, each robot first constructs its local context, then independently infers an action, and finally both actions are committed to the environment simultaneously. 


This keeps decision making decentralized while ensuring synchronized interaction for evaluation. In real two-machine deployment, each robot runs the same local pipeline and exchanges only semantic packets. Messages may arrive asynchronously, so each robot only uses partner semantics within a bounded freshness window. 


Consequently, temporary delay degrades coordination quality but does not block local progress. Together, SVB, EDR, and SPE directly address the two core problems in collaborative VLN: static coordination and non-synchronized evaluation.

\section{Experiments}
\label{sec:experiments}

\begin{figure*}[!t]
  \centering
  \includegraphics[width=\textwidth]{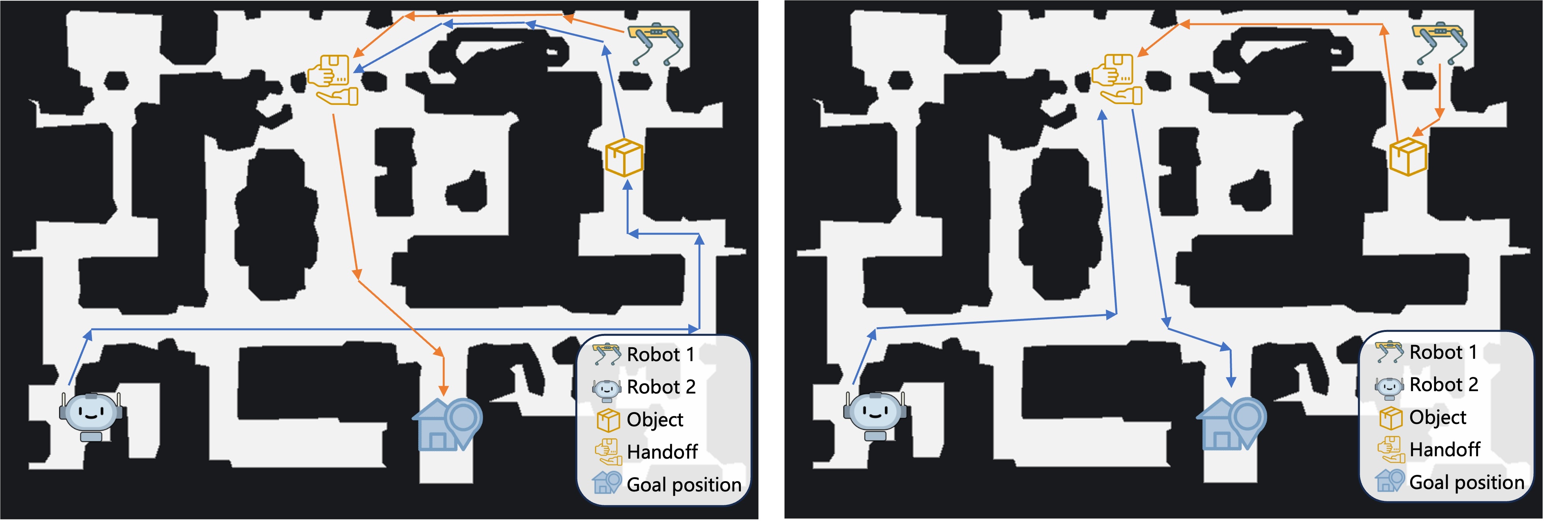}
  \caption{Illustration of dynamic subtask reassignment in DeCoNav. 
  \textbf{Left:} Initial task allocation, where Robot\,1  navigates to the object and proceeds to the handoff point,  and Robot\,2 moves from the handoff point to the goal. 
  \textbf{Right:} After event-triggered dialogue replanning, 
  subtasks are reassigned based on updated semantic evidence, 
  resulting in shorter overall paths for both robots.}
  \label{fig:3-2}
\end{figure*}

\subsection{Benchmark Experiments}
\label{subsec:bench_experiments}

\subsubsection{Data Statistics}
\label{sec:data_statistics}

\begin{table}[!t]
\centering
\footnotesize
\caption{CoNavBench vs.\ DeCoNavBench; $\pm$ denotes standard deviation.}
\label{tab:conav_vs_deconav}
\begin{tabular}{lcc}
\toprule
\textbf{Metric} & \textbf{CoNavBench~\cite{wang2026conavbench}} & \textbf{DeCoNavBench} \\
\midrule
Tasks & 992 & 1{,}213 \\
HM3D scenes & 128 & 176 \\
Mean steps/robot & $61.2 \pm 36.5$ & $139.9 \pm 62.3$ \\
Mean path/robot (m) & $9.7 \pm 6.7$ & $20.6 \pm 10.4$ \\
Combined path (m) & $19.4 \pm 11.0$ & $41.2 \pm 17.8$ \\
\bottomrule
\end{tabular}
\end{table}

The released benchmark comprises 1{,}213 collaborative navigation tasks across 176 HM3D scenes, with an average path length of 20.6 metres per robot. Each task is stored with pre-rendered RGB-D observations for both robots, totalling over 2.05 million images consisting of 3 RGB and 3 depth views per step per robot.

\subsubsection{Room-Type Semantic Alignment (RTSA) Evaluation}
\label{subsec:room_recon_eval}
This section reports quantitative results of the RTSA pipeline.

\noindent\textbf{Stage 1: Rule-Based Keyword Inference.}
The keyword-only stage achieves 64.1\% accuracy against the final human-verified ground truth, with high precision for distinctive rooms (\textit{Bathroom}: 87.6\%, \textit{Bedroom}: 84.5\%) but poor recall for ambiguous categories (\textit{Corridor}: 36.1\%, \textit{Dining Room}: 19.3\%).

\noindent\textbf{Stage 2: VLM Panoramic Classification + Cross-Model Verification.}
Each room panorama is first classified by GPT and then verified with Qwen-VL when disagreements occur. Used independently, GPT achieves 88.5\% accuracy and Qwen-VL achieves 85.0\%; combining both models with the rule-based keywords through a three-way majority vote substantially reduces noise, yielding 92.1\% overall accuracy across all 2{,}469 rooms.

\noindent\textbf{Stage 3: Manual Review.}
All remaining uncertain or conflicting cases are manually adjudicated, improving label quality from the automated 92.1\% to 100.0\% accuracy on all 2{,}469 rooms.

Table~\ref{tab:room-label-quality} compares room-type label quality
across three datasets on the same 181 HM3D scenes. We report two
metrics: \emph{coverage}, the proportion of rooms assigned a
recognisable room-type label, and \emph{correctness}, the proportion
of labelled rooms whose label matches the human-verified ground truth.
LHPR-VLN covers only 73\% of rooms with 61\% correctness, and
CoNavBench improves coverage to 90\% but still has 32\% of its labels
incorrect. In contrast, our three-stage RTSA pipeline labels all
2{,}469 rooms with 100\% verified correctness, eliminating the
annotation gaps that could otherwise introduce noise into downstream
navigation training and evaluation.

\begin{table}[!t]
  \centering
  \caption{Room-type label quality across three datasets (181 HM3D scenes, 2{,}469 rooms).}
\label{tab:room-label-quality}
\footnotesize
\resizebox{\columnwidth}{!}{
\begin{tabular}{l c c c}
\toprule
\textbf{Metric} & \textbf{LHPR-VLN}~\cite{song2025towards} & \textbf{CoNavBench}~\cite{wang2026conavbench} & \textbf{DeCoNavBench (ours)} \\
\midrule
Labelled rooms       & 1{,}803          & 1{,}582          & 2{,}469 \\
Unlabelled rooms     & 666              & 168              & 0 \\
Total rooms          & 2{,}469          & 1{,}750          & 2{,}469 \\
\midrule
Coverage        & 73\%               & 90\%               & 100\% \\
Correctness     & 61\%               & 68\%               & 100\% \\
\bottomrule
\end{tabular}
}
\end{table}

\subsection{Simulation Experiments}
\label{subsec:sim_experiments}

\subsubsection{Experimental Setup}
\label{subsubsec:experimental_setup}
We evaluate dual-robot task performance on DeCoNavBench.
The main comparison is between two trained policies:
\textbf{(i)} a CoNavBench-trained baseline model and
\textbf{(ii)} a DeCoNav-trained model. Both models are tested on the
same test split. These test episodes are held out during
training and are never used for model fitting.

Each episode runs Fetch and Spot concurrently in one shared simulator.
The action space is
\{\texttt{move\_forward} (0.25\,m), \texttt{turn\_left} (15\textdegree),
\texttt{turn\_right} (15\textdegree), \texttt{stop}\}.
Rollout follows SPE for synchronized dual-agent execution. We report
BSR and per-robot SR, SPL, ISR, and NE.

\subsubsection{CoNavBench vs.\ DeCoNav}
\label{subsubsec:conavbench_vs_ours}
Table~\ref{tab:conavbench_vs_ours} compares a CoNavBench baseline
model and a DeCoNav model. During evaluation, the DeCoNav model
runs with the full DeCoNav method (SVB+EDR) under synchronized
execution. Both models are evaluated on the same test split.
\begin{table}[!t]
\centering
\footnotesize
\caption{Comparison between CoNavBench and DeCoNav on the same test split.}
\label{tab:conavbench_vs_ours}
\resizebox{\columnwidth}{!}{
\begin{tabular}{lccccc}
\toprule
\textbf{Setting} & \textbf{SR$\uparrow$} & \textbf{BSR$\uparrow$} & \textbf{ISR$\uparrow$} & \textbf{SPL$\uparrow$} & \textbf{NE$\downarrow$} \\
\midrule
CoNavBench baseline & 0.28 & 0.13 & 0.35 & 0.18 & 5.19 \\
DeCoNav (Ours) & 0.39 & 0.22 & 0.47 & 0.32 & 4.75 \\
\bottomrule
\end{tabular}
}
\end{table}
Compared with CoNavBench, DeCoNav achieves an absolute SR improvement of 0.11 (39.3\% relative), indicating substantial system-level benefit from the proposed collaborative framework.

\subsubsection{Ablation on DeCoNav}
\label{subsubsec:ablation_deconav}
Table~\ref{tab:ablation_deconav} isolates the effect of adding DeCoNav under the same synchronized protocol. \textbf{w/o DeCoNav} disables dialogue-semantic updates, reducing coordination to a mostly static behavior, and \textbf{w/ DeCoNav} enables SVB+EDR for online dynamic replanning. Unlike the cross-framework comparison in Section~\ref{subsubsec:conavbench_vs_ours}, this ablation holds the rollout protocol constant and varies only the coordination mechanism, thereby isolating the contribution of event-triggered semantic coordination.

\begin{table}[!t]
  \centering
  \footnotesize
  \caption{Ablation of DeCoNav on the same test split.}
  \label{tab:ablation_deconav}
  \resizebox{\columnwidth}{!}{
    \begin{tabular}{lccccc}
      \toprule
      \textbf{Setting} & \textbf{SR$\uparrow$} & \textbf{BSR$\uparrow$} & \textbf{ISR$\uparrow$} & \textbf{SPL$\uparrow$} & \textbf{NE$\downarrow$} \\
      \midrule
      w/o DeCoNav & 0.32 & 0.15 & 0.41 & 0.24 & 4.93 \\
      w/ DeCoNav  & 0.39 & 0.22 & 0.47 & 0.32 & 4.75 \\
      \bottomrule
    \end{tabular}
  }
\end{table}


\begin{figure*}[!t]
  \centering
  \includegraphics[width=\textwidth]{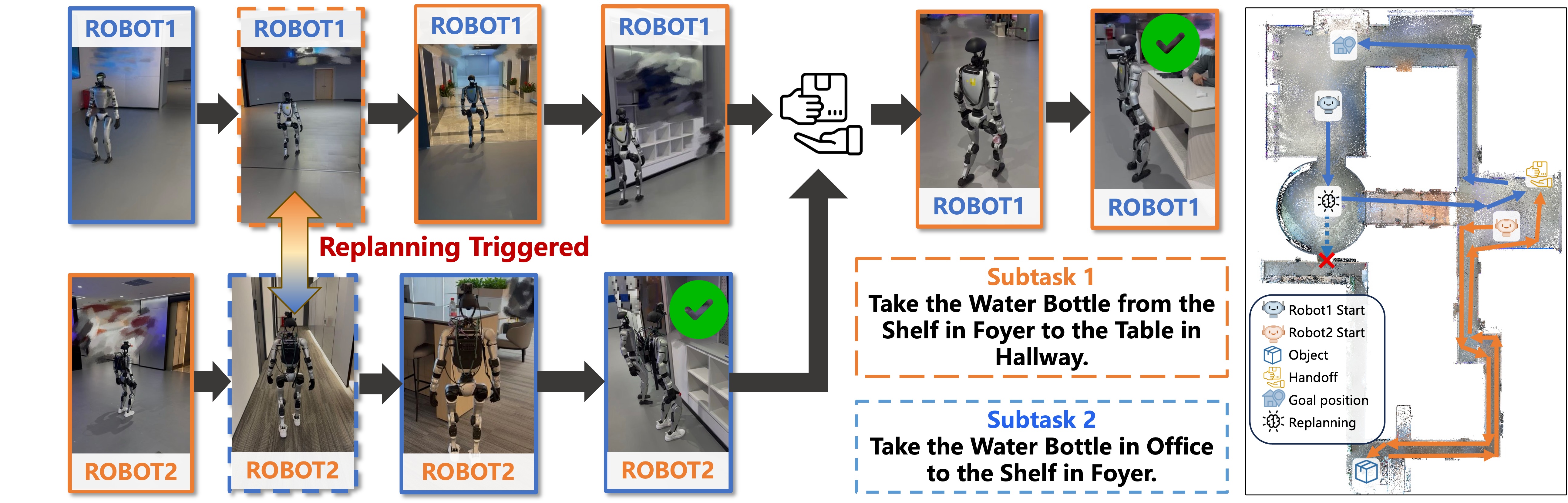}
  \caption{Real-robot deployment on a collaborative object transport
  task. Robot\,1 and Robot\,2 operate as equal peers and execute
  decomposed subtasks concurrently under mapless exploration. When
  Robot\,1 discovers a previously passable corridor is locked, the EDR module triggers online replanning. Through
  ROS2 communication and shared semantic memory, both
  robots negotiate a subtask swap that yields shorter overall paths, and
  complete the task.}
  \label{fig:real-robot}
\end{figure*}

\subsection{Real-Robot Experiments}
\label{subsec:real_robot_experiments}

To validate the practical applicability of DeCoNav beyond simulation, we
deploy the framework on a pair of Unitree humanoid robots in an
unstructured indoor office environment. Both robots operate as equal
peers without a centralised coordinator, communicating via ROS2 topics to
share observed room types and object locations accumulated during
mapless exploration. The system executes a representative collaborative
transport task end-to-end: \textit{``Take the water bottle in the office
to the table in the hallway.''} The task planner decomposes this
instruction into two concurrent subtasks (Fig.~\ref{fig:real-robot}):
Robot\,2 navigates to the office, picks up the water bottle, and
delivers it to the shelf in the foyer as a handover point; Robot\,1
then retrieves the bottle from the shelf and transports it to the
designated table in the hallway.

During execution, Robot\,1 encounters an unexpected obstacle: a
previously traversable corridor is locked and impassable.
Because the blocked path substantially increases Robot\,1's estimated
cost for its current subtask, the EDR module is triggered. The two
robots exchange their real-time positions and
historical semantic memory
through ROS2 communication. Based on this shared
information, both robots jointly determine that, given their respective
current locations, swapping subtasks yields shorter overall paths.
They accordingly reassign their roles, as
shown in Fig.~\ref{fig:real-robot}. After
the updated allocation, both robots proceed along more efficient routes and complete the collaborative handover within a single
continuous episode without manual intervention. This case demonstrates a distinctive advantage of DeCoNav: peer robots
can autonomously negotiate subtask reassignment through real-time
communication and shared semantic memory, jointly optimising navigation distance and task
completion time in dynamic environment.

\section{Conclusion}
This paper introduces DeCoNav and DeCoNavBench for long-horizon collaborative VLN.
DeCoNav is a decentralized method that combines semantic-state communication,
event-triggered online replanning, and synchronized dual-agent execution.
The same method formulation is consistent with two-machine deployment
through asynchronous semantic message passing and freshness-bounded
context fusion.
DeCoNavBench, built by extending CoNavBench with the ROVE pipeline,
provides room- and object-verified episodes for evaluating this setting at scale.
Our results demonstrate that, beyond the benchmark foundation laid by CoNavBench, advancing collaborative long-horizon VLN demands both synchronized dual-robot evaluation and adaptive online coordination. We hope DeCoNav and DeCoNavBench will serve as a stepping stone toward this goal.

\bibliographystyle{IEEEtran}
\bibliography{references}

\end{document}